# Single and Multi-Hop Question-Answering Datasets for Reticular Chemistry with GPT-4-Turbo

Nakul Rampal[1,2,3], Kaiyu Wang[1,2], Matthew Burigana[1,2], Lingxiang Hou[1,2], Juri Al-Johani[3,4], Anna Sackmann[5], Hanan S. Murayshid[8], Walaa Abdullah Al-Sumari[8], Arwa M. Al-Abdulkarim[8], Nahla Eid Al-Hazmi[9,10], Majed O. Al-Awad[8], Christian Borgs[3,4,*], Jennifer T. Chayes[3,4,6,7,*], Omar M. Yaghi[1,2,3,10,*]

[1]Department of Chemistry, University of California, Berkeley, California 94720, United States

[2]Kavli Energy Nanoscience Institute, University of California, Berkeley, California 94720, United States

[3]Bakar Institute of Digital Materials for the Planet, College of Computing, Data Science, and Society, University of California, Berkeley, California 94720, United States

[4]Department of Electrical Engineering and Computer Sciences, University of California, Berkeley, California 94720, United States

[5]Data Services Librarian, University of California, Berkeley Library, United States

[6]Department of Mathematics, University of California, Berkeley, California 94720, United States

[7]Department of Statistics, University of California, Berkeley, California 94720, United States

[8]Artificial Intelligence & Robotics Institute, Economies of the Future Sector, King Abdulaziz City for Science and Technology (KACST), Riyadh 12354, Saudi Arabia

[9]Hydrogen Technologies Institute, King Abdulaziz City for Science and Technology, P.O. Box 6086, Riyadh 11442, Saudi Arabia

[10]KACST−UC Berkeley Center of Excellence for Nanomaterials for Clean Energy Applications, King Abdulaziz City for Science and Technology, Riyadh 11442, Saudi Arabia

*Email: borgs@berkeley.edu, jchayes@berkeley.edu, yaghi@berkeley.edu

**Abstract:** The rapid advancement in artificial intelligence and natural language processing has led to the development of large-scale datasets aimed at benchmarking the performance of machine learning models. Herein, we introduce 'RetChemQA,' a comprehensive benchmark dataset designed to evaluate the capabilities of such models in the domain of reticular chemistry. This dataset includes both single-hop and multi-hop question-answer pairs, encompassing approximately 45,000 Q&As for each type. The questions have been extracted from an extensive corpus of literature containing about 2,530 research papers from publishers including NAS, ACS, RSC, Elsevier, and Nature Publishing Group, among others. The dataset has been generated using OpenAI's GPT-4 Turbo, a cutting-edge model known for its exceptional language understanding and generation capabilities. In addition to the Q&A dataset, we also release a dataset of synthesis conditions extracted from the corpus of literature used in this study. The aim of RetChemQA is to provide a robust platform for the development and evaluation of advanced machine learning algorithms, particularly for the reticular chemistry community. The dataset is structured to reflect the complexities and nuances of real-world scientific discourse, thereby enabling nuanced performance assessments across a variety of tasks. The dataset is available at the following link: https://github.com/nakulrampal/RetChemQA

**TOC Entry:**

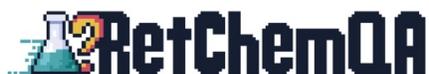

**Main:**

Given the increasing application of large language models (LLMs) across various scientific domains, including chemistry, the development of benchmark datasets for evaluating their performance is crucial. While benchmark datasets for many tasks across different subjects already exist—such as PubMedQA [1] for biomedical questions, HotPotQA [2] for complex question answering, and SQuAD [3], [4] for reading comprehension—there is a noticeable lack of datasets for tasks specific to reticular chemistry. This study aims to bridge this gap by introducing a question and answer (Q&A) dataset, which we have named RetChemQA, tailored to the unique demands of reticular chemistry [5], [6]. While for the researchers working at the intersection of computer science and reticular chemistry, this dataset provides a standard against which new LLMs and methodologies can be benchmarked, allowing for the development of new algorithms, and hardware, we believe this work also holds importance for those who have limited or no background in computer science. For reticular chemists working in the wet lab, the development of models/methodologies that provide more accurate predictions and hypotheses, will in turn lead to a reduction in the time and resources required for empirical testing in the lab. Additionally, since the output provided by an LLM is strongly correlated with the nature of the prompt provided [7], it would be useful to have a method where the prompt optimization is automated. This avoids the (i) inconsistency that might result from variation of a prompt seeking a particular output and (ii) subjective human judgement leading to arduous, time-consuming, and impractical approaches to evaluate LLM outputs and prompts for tasks that require thousands of repeating LLM calls.

The introduction of the RetChemQA dataset will enable the development and application of automated prompt optimization and evaluation frameworks like DSPy [8], which can be useful when using LLMs for tasks such as refining experimental designs and optimizing synthesis conditions. Furthermore, building chat-based user interfaces for these prompt optimization frameworks will further lower the barriers to entry for scientists with limited or no knowledge of computer science.

In this contribution, we report RetChemQA, a dataset containing a total of about ~90,000 Q&As automatically generated using GPT-4-Turbo from a corpus of about 2,530 papers including both the manuscript (MS), and supplementary information (SI), where available. The dataset contains questions of 3 types: (i) *Factual questions*: these type of questions are those for which an answer is a stated fact, (ii) *Reasoning questions*: these type of questions require an understanding of the text to answer, and (iii) *True/False questions*: these are categorical questions that have a True/False answer. An example of each type of question is shown in **Figure 1**.

**Question Types**

| Output | Output | Output |
|---|---|---|
| **"Question"**: "What is the role of anhydrous $Zn(O_3SCF_3)_2$ in the synthesis of ZIF-100?"<br><br>**"Answer"**: "Anhydrous $Zn(O_3SCF_3)_2$ is used in place of $Zn(NO_3)_2 \cdot 4H_2O$ to achieve less hydrated conditions, promoting the crystallization of ZIF-100"<br><br>**"Difficulty"**: "Medium"<br><br>**"Type"**: "Factual" | **"Question"**: "Why are ZIFs considered promising materials for carbon dioxide capture?"<br><br>**"Answer"**: "ZIFs are considered promising materials for carbon dioxide capture due to their selective adsorption of $CO_2$, high thermal and chemical stability, and ease of fabrication."<br><br>**"Difficulty"**: "Medium"<br><br>**"Type"**: "Reasoning" | **"Question"**: "True or False: ZIF-100 has a larger primitive cell volume than MIL-101"<br><br>**"Answer"**: "True"<br><br>**"Difficulty"**: "Easy"<br><br>**"Type"**: "True or False" |
| **Factual**<br>A question for which the answer is a stated fact. | **Reasoning**<br>A question that requires an 'understanding' of the text to answer. | **True/False**<br>A question that for which the answer is either *True* or *False*. |

**Figure 1: Type of questions in RetChemQA.** The dataset consists of three main types of question, from left to right, (i) Factual, (ii) Reasoning, and (iii) True/False. In the example shown above, the questions have been generated using GPT-4-Turbo using the prompt shown further below (**Figure 4**) from the following DOI: 10.1038/nature06900 [9].

Moreover, the Q&A pairs generated are also categorized based on the difficulty levels: Easy, Medium, and Hard. Building on the categorization framework further, questions can also be classified on the number of reasoning steps required to answer them. Questions that require a single step of reasoning are termed as single-hop questions, and those that require multiple steps of reasoning, are termed as multi-hop questions. When working with the scientific literature, a single-hop question can often be answered by consulting only a single sentence provided in the MS or SI. On the other hand, a multi-hop question, will often require information from multiple places in the MS and SI to answer. Examples of both the single-hop and multi-hop question types are shown in **Figure 2**. For the single-hop example question: "At what temperature range was the solvent-exchanged and evacuated ZIF-11 heated for gas-sorption analysis preparation?" we see that the answer generated is from a single contiguous piece of text taken from the MS, while for the multi-hop example question: "What temperature range was used for the solvothermal synthesis of ZIFs?" we see that the answer generated "The solvothermal synthesis was carried out at temperatures ranging from 85-150 °C" includes information from multiple parts of both the MS and SI of the paper. Interestingly, if the question were to be answered as a single-hop question, the question would have probably been answered incorrectly, as in the MS under the section "Typical ZIF" synthesis the temperature given is "140 °C" while in the SI, where the individual synthesis conditions of each ZIF are provided, the temperature mentioned is different for each ZIF, so any answer generated would have not included a range of temperatures as this information is not explicitly given anywhere in the paper.

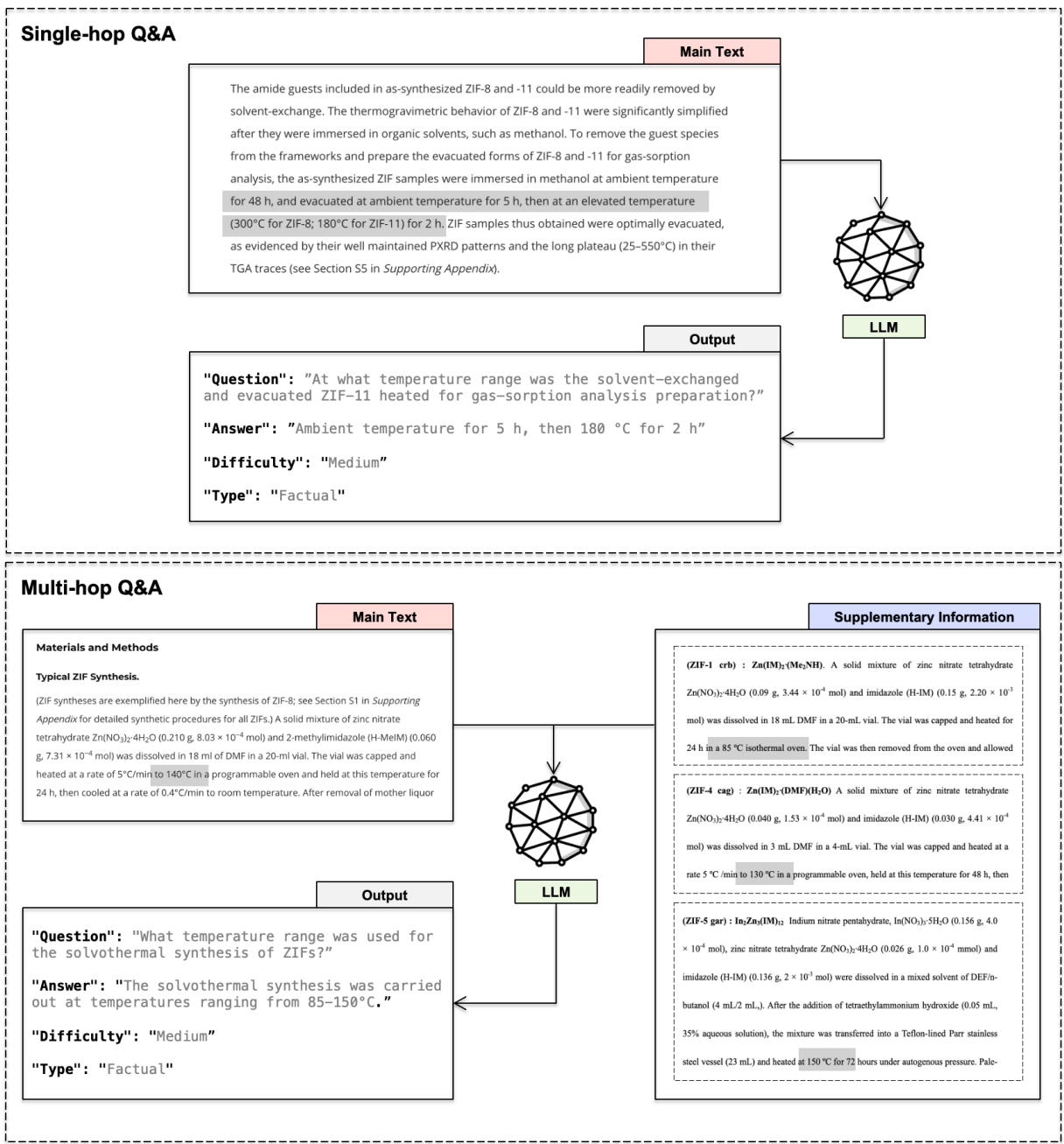

**Figure 2: Single-hop *vs*. Multi-hop Q&A.** Flow diagram of the information in a single-hop (top) and multi-hop (bottom) Q&A generation task. A single-hop Q&A is defined as one that requires only a single step of reasoning to answer; often this involves retrieving information from a single sentence of a given paper. A multi-hop Q&A is defined as one that requires multiple steps of reasoning to answer; often this involves retrieving information from multiple different parts of a MS. In the example shown, data must be collected from both the MS and SI to answer the question. In the example shown above, the questions have been generated using GPT-4-Turbo (gpt-4-0125-preview) using (i) the prompt shown in **Figure 4** (*left*) for the single hop Q&A, and (ii) the prompt shown in **Figure 4** (*right*) for the multi-hop Q&A, from the DOI: 10.1073/pnas.0602439103 [10].

**Methods:**

To build a corpus of text, we started with the CSD MOF Subset (April 2023) [11] that contains information of about 122,738 MOFs in 51,046 DOIs. Of the 122,738 MOFs present in the subset, we found that 8,089 MOF entries did not have an associated DOI — these MOF entries were removed. Next, we decided to consider only mainstream publishers: RSC, ACS, Wiley, Elsevier, AAAS, AIP, APS, Beilstein, CCS, De Gruyter, Frontiers, IOP, IUCr, NAS, Nature, Royal Society Publishing, T&F, University Press (Oxford, Cambridge, Tsinghua). Full names for the acronyms are given in the SI in Table S1. After applying this criterion, we had 49,044 DOIs to work with. Finally, we further narrowed our corpus of text by working with only specific journals for each publisher (For more details, please see **Table S2**). In total, 2,530 DOIs were processed: Nature (220) RSC (215), Elsevier (82), ACS (1,283), AAAS (46), Wiley (653), NAS (10), CCS (10), AIP (5), and APS (6). For each publisher, all the text and data mining were performed keeping in mind the contractual agreements the University of California, Berkeley Library has with the individual publishers. To generate the Q&A + synthesis conditions datasets, the latest model from OpenAI, GPT-4-Turbo (*gpt-4-0125-preview*) was used. In total, 337,577,236 tokens were processed, at a cost of $3,600 for the whole project (including development and testing). The cost of generating a dataset (Q&A or synthesis conditions) is ~$1,000; this translates to a cost of $0.40 per DOI.

      We started with generating the single-hop Q&A dataset according to the workflow shown in **Figure 3**. To begin with, the processing environment is initialized. Next, for each *document_dir* given in each *publisher_dir*, the files are parsed and the combined text is then tokenized and then passed to the LLM for processing. A more detailed description data processing workflow algorithm is given in **Figure S2**. In the prompt provided, we explicitly specified that (i) the total of number of Q&As we want, in this case 20, and (ii) the number of different question types we want: 6 Factual, 7 True/False, and 7 Reasoning. We also mentioned the labels we want to include in the dataset: the question, the answer, the difficulty level, and the type of question. Here, a deliberate attempt was made to strike the right balance in that the prompt needed to be sufficiently open-ended to encourage creativity, yet sufficiently structured to provide clear direction. To generate the multi-hop Q&A dataset, we initially did a simple modification to the prompt used to generate the single-hop Q&A dataset. We replaced the word 'single-hop' with 'multi-hop' in the whole prompt. Interestingly, this did not significantly change the output generated. There were many instances where both the number and type of single-hop and multi-hop questions generated for each DOI were very similar. Our goal therefore was then to develop a prompt that would diversify the types of questions generated. By providing additional context and including details like: "A multi-hop Q&A is one that requires multi step reasoning to come to an answer (this information can come from any part of the paper, both MS and SI). To give you more details: A multi-hop Q&A will always involve going through multiple parts of the paper to come to an answer. This may include different paragraphs, different pages, and also

different documents (i.e. the MS and SI)" in the prompt, we were able to significantly reduce but not completely eliminate the similar question types. The final prompts used to generate the single-hop and multi-hop Q&A datasets are shown in **Figure 4**. The generation of synthesis conditions is far from trivial given that each paper uses different variables and formats for presenting information. This variability makes it difficult to establish a fixed set of variables to provide to the LLM. Moreover, each paper may contain synthesis conditions for more than one material, with the maximum number of materials unknown. Given these uncertainties, devising the right prompt was challenging. Initial attempts were made to keep the prompt as open-ended as possible, allowing the LLM to decide on the number and type of variables it deemed necessary. However, without sampling all the text provided in the dataset, it would be impossible to identify all the variables. Keeping the prompt open-ended resulted in the extraction of a lot of unnecessary information, including experimental characterization data such as crystal structure data and NMR peaks, which is often included under the 'synthesis conditions' section of a paper. To minimize this, we explicitly added a statement to the prompt instructing the LLM not to extract any experimental characterization data. Although this approach significantly improved the nature of the outputs generated, it could not completely eliminate the extraction of experimental characterization data. It is important to note that for DOIs associated with Wiley, we could only process the MS and not the SI, as we could not automate the downloading of the SI files. The final prompt used to generate the synthesis conditions dataset is shown in **Figure S1**.

For the multi-hop dataset, the data processing failed for 56 DOIs while for the single-hop dataset, the data processing failed for 34 DOIs. For the single-hop dataset, we came across an example (DOI:10.1021/jacs.3073512) where the .json file mentioned "Add more questions as needed" – this was classified as an incomplete/failed generation. For some, the output generated was not in the format of a .json file and such outcomes were counted as a failed DOI also. A summary of the errors/outputs generated for each of the DOIs for both the multi-hop and single-hop datasets is given in the supplementary files. For the synthesis conditions dataset, the data processing failed for 62 DOIs. A summary of the errors and the outputs generated for these 62 DOIs is given in the supplementary files.

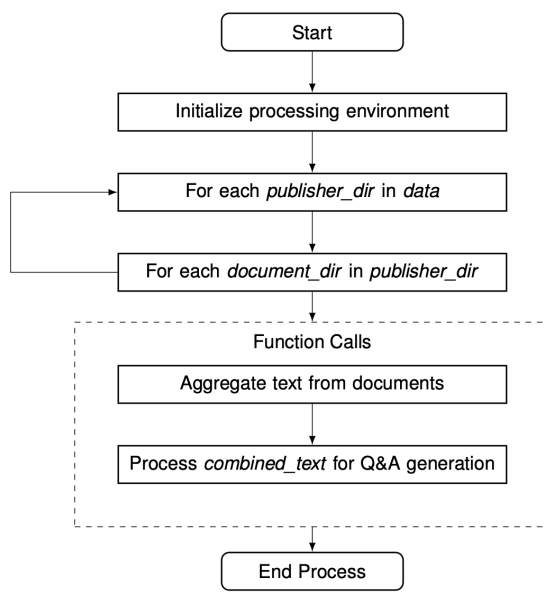

**Figure 3: Dataset generation workflow.** The figure depicts the dataset generation workflow with arrows indicating the order in which the steps are performed. For each *publisher_dir* (Nature, RSC, Elsevier, ACS, AAAS, Wiley, NAS, CCS, AIP, and APS) and every *document_dir* in a given *publisher_dir* the following function calls are made: 1. Aggregation of text from the documents (both the MS and SI, where available) followed by processing of the *combined_text* using the LLM, in this case GPT-4-Turbo, to generate the Q&A pairs. A more detailed dataset generation workflow/algorithm is given in **Figure S2**.

In addition, for both the Q&A (single-hop and multi-hop) and synthesis conditions datasets we have omitted the 'context' label that is often included to avoid any copy right issues or concerns with the publishers. We would recommend the readers to use the entire text of a particular DOI including SI, where available, as the 'context' for a given Q&A. Importantly, each file is named as follows: [DOI]_single-hop.json or [DOI]_multi-hop.json or [DOI]_synthesis-conditions.json, this should make working with the dataset easier as each file will only contain information specific to the DOI specified in the prefix of the filename.

**Single-hop Q&A**

**Prompt**

```
"System": " You are a single hop Question and
Answering (Q&A) dataset generation agent. A single
hop question and answer set is one that requires a
single step of reasoning. You are required to go
through the given text and identify the synthesis
conditions and based on those synthesis conditions
develop a set of 20 Q&As. There may be information
about the synthesis conditions of more than one
material in the text. For example, you may come
across a series of different materials such as ZIF-1,
ZIF-2, .... ZIF-12. Please try to diversify the types
of questions that you include. Please also try to
include a question for each material you come across
in the paper. Please feel free to include labels that
are also used in some of the most widely used Q&A
datasets e.g., the question, the answer, the
difficulty level, and the type of question. the
different types of questions are factual, reasoning
(single step reasoning), and True or False. Please
generate 6 'factual' type questions, 7 'reasoning'
type questions, and 7 True or False type questions."

"User": " Generate a single hop .json file for the
following text. Please include questions of different
types including factual (6 questions), single-step
reasoning (7 questions), and True or False (7
questions) : {combined_text}."
```

**Multi-hop Q&A**

**Prompt**

```
"System": " You are a multi-hop Question and
Answering (Q&A) dataset generation agent. A multi-hop
Q&A is one that requires multi step reasoning to come
to an answer (this information can come from any part
of the paper, both MS and SI). To give you more
details: A multi-hop Q&A will always involve going
through multiple parts of the paper to come to an
answer. This may include different paragraphs,
different pages, and also different documents (i.e.
the manuscript and the supplementary information).
You are required to go through the given text and
identify the synthesis conditions and based on those
synthesis conditions develop a set of multi-hop
(questions that require multiple steps of reasoning)
20 Q&As for each DOI. There may be information about
the synthesis conditions of more than one material in
the text. For example, you may come across a series
of different materials such as ZIF-1, ZIF-2, ....
ZIF-12. Please diversify the type of questions to
encompass different ideas and materials. Please feel
free to include labels that are also used in some of
the most widely used Q&A dataset for e.g., the
question, the answer, the difficulty level, and the
type of question. the different types of questions
are factual, reasoning (single step reasoning), and
True or False. Please generate 6 'factual' type
questions, 7 'reasoning' type questions, and 7 True
or False type questions. For factual questions,
please try to be creative with the questions as it
should require information from different parts of
the text to answer"

"User": "Generate a multi-hop Q&A json file for the
following text. Please include questions of different
types including factual (6 questions), single-step
reasoning (7 questions), and True or False (7
questions): {combined_text}."
```

**Figure 4. Prompts used to generate the set of Q&As.** The prompt used to generate the single-hop Q&As is shown on the left, and the prompt used to generate the multi-hop Q&As is shown on the right. Each prompt consists of messages that are adopted to specific 'roles' to guide the model's response. The "system" role provides the high-level instructions, while the "user" role provides the query. The "combined_text" variable holds all the text information contained in the MS and SI (where available). This information is provided as part of the prompt to GPT-4-Turbo.

Existing Q&A dataset evaluation criterion such as accuracy, precision, etc. are based on the premise that the question for which the answer is being evaluated is in itself correct. This may not always hold true when the set of question-answer pairs is generated using an LLM. It is important to keep in mind that LLMs may also 'hallucinate questions'— generate a question answer pair from information *not* provided as 'context' in the prompt to the LLM —and therefore the answer to that question, may also be incorrect. Hence, it was required that we come up with an evaluation metric that takes into account such outcomes. In addition, existing evaluation frameworks in literature do not generalize well across different question types. For example, for a Factual/Reasoning question there is no 'negative' answer and therefore classifying an answer type as 'False Negative (FN)' is not possible. This required the development of a new evaluation framework, that although similar to the framework used previously in literature, is tuned for our particular Q&A generation approach.

In the evaluation approach considered in this paper, the question-answer pair is first assessed based on whether it has been generated from the context provided in the prompt or not. If the question has been generated from the context provided, we next evaluate whether the answer to the question is correct or not. If the answer to the question is correct, the question answer pair is classified as 'True Positive (TP)'; else, the question answer pair is classified as 'False Positive (FP)'. On the other hand, if the question-answer pair generated is *not* from the context provided to the LLM and the answer to the question generated is correct, for example: The answer to a hallucinated question is "I cannot answer this question from the information provided in the context/prompt'— The LLM has itself identified that this is an hallucinated question, it is classified as 'True Negative (TN)'; else the question answer pair is classified as 'FN'. The evaluation framework described above is also shown as a flowchart in **Figure S3**. Moreover, the evaluation framework can also handle question-answer pairs that have been incorrectly classified by the LLM: for example, a Reasoning/Factual question has been classified as a True/False question; the question-answer pair is classified as 'out of context', allowing us to penalize the LLM for the incorrect categorization of the question-answer pair. We introduce a similar penalty if the LLM generates a single-hop question-answer pair when in the prompt provided it is explicitly stated to generate a multi-hop question-answer pair. This evaluation framework is broadly applicable to all the different question types considered in this dataset and therefore allows for comparison of the performance of the LLM in the generation of the different question types. Examples of question-answer pairs classified as TP, FP, TN, and FN in the single-hop and multi-hop datasets are shown in **Figure S4** and **S5**. Following the classification of each Q&A pair, the performance of the LLM is assessed based on the following metrics, and is specific to the evaluation framework described above:

1. Accuracy: This is a measure of the ability of the LLM to correctly answer questions that have been generated both in or out of context — here, a penalty is introduced for answers that are *incomplete* or wrong whether the question is in or out of context. It is defined as the ratio of the sum of the correctly answered questions, (TP + TN) to the total number of possible outcomes (TP + TN + FP + FN). A high accuracy score indicates better performance while a low accuracy indicates otherwise.
2. Precision: This is a measure of the ability of the LLM to accurately answer questions that have been generated only in context — In addition to the penalties introduced above, here, a penalty is also introduced for (i) hallucinated questions even if answered correctly, (ii) incorrectly generated questions, and (iii) incorrectly categorized questions. It is defined as the ratio of accurately answered in context questions (TP) to the total number of possible outcomes (TP + FN + FP + FN). A high precision score is desired as it indicates better performance; a low precision score indicates otherwise.

3. Hallucination Rate: This is a measure of proportion of Q&A pairs hallucinated by the LLM. It is defined as the ratio of the sum of hallucinated Q&A pairs (TN + FN) to the total number of possible outcomes (TP + TN + FP + FN). A low hallucination rate indicates better performance, while a high hallucation rate indicates otherwise.
4. Hallucination Capture Rate: This is a measure of the LLM's ability to identify and correct a hallucinated (out-of-context) question it has generated itself. It is defined as the ratio of hallucinated questions generated but answered correctly (TN) to the total number of hallucinated questions generated (TN + FN). A high hallucination capture rate is desired as it means that the LLM is able to identify its mistake, while a low hallucination capture rate indicates otherwise.

As discussed previously, the evaluation of the synthesis conditions dataset is far less straightforward. This is because for each DOI, the number of variables and the format in which the synthesis conditions are reported is different. In addition, no one single set of synthesis conditions follows a standard format. This complicates the development of a performance assessment metric that encompasses all the different synthesis conditions. Therefore, we propose a binary Yes (Y) or No (N) assessment metric. If for a particular material, the entire set of the synthesis conditions extracted is correct, we assign a '**Y**'; if it is incorrect or incomplete we assign an '**N**'– this is criterion 1. If the conditions *do not* include experimental characterization data, we assign an '**N**' otherwise we assign it a '**Y**'— this is criterion 2. To assess the performance of the LLM in extracting synthesis conditions we then calculate the ratio of the number of **Y**s generated to the total number of outcomes (no. of **N**s + no. of **Y**s) for criterion 1 and the ratio of the number of **N**s generated to the total number of outcomes (no. of **N**s + no. of **Y**s) for criterion 2. A product of the two ratios then allows us to determine the performance of the LLM in extracting the synthesis conditions — we term this the 'Obedience' score. This is because a high 'Obedience' score would imply that the LLM is strictly 'obeying' the two main instructions provided in the prompt: (i) extract all the synthesis conditions for a given a material, and (ii) to not include any experimental characterization data. On the other hand, a low 'Obedience' score would mean that the LLM fails to adequately follow the instructions provided in the prompt.

**Results and Discussion:**

In total 89,551 question-answer pairs were generated with 54 % (48,454) being single-hop Q&As and 46 % (41,097) multi-hop Q&As. For both the single-hop (2,496 DOIs) and multi-hop (2,474 DOIs) datasets an approximately equal number of DOIs were processed. The distribution of the different question types in both the single-hop and multi-hop datasets is shown in **Figure 5**. For the single-hop Q&A dataset, for each DOI, ~ 19 questions are generated of which on average 7 are Factual, 7 are True/False and 6 are Reasoning.

On the other hand, for the multi-hop Q&A dataset, for each DOI, ~ 17 questions are generated of which on average 5 are Factual, 6 are True/False, and 5 are Reasoning. A detailed summary of the distribution of the different question types, along with their difficulty levels is given in **Table 1**.

**Figure 5. Results for the RetChemQA dataset.** Total number of questions generated (and the type: Factual, Reasoning, and True/False) in the single-hop and multi-hop datasets(top); and a word cloud showing the different synthesis related keywords generated in the synthesis conditions dataset (bottom). The size of a word indicates its frequency or importance in the text being analyzed, where larger words represent those that appear more frequently, while smaller words represent those that appear less frequently.

**Table 1: Average number of different question types (Factual, True/False, Reasoning) and difficulty levels (Easy, Medium or Hard) generated per DOI.** For the single-hop Q&A dataset, a total of 48,454 question-answer pairs were successfully generated from 2,496 DOIs. For the multi-hop Q&A dataset, a total of 41,097 question-answer pairs were successfully generated from 2,474 DOIs. Each value in the table has been rounded to the nearest whole number.

|  | Factual | | | True/False | | | Reasoning | | | Total |
| --- | --- | --- | --- | --- | --- | --- | --- | --- | --- | --- |
|  | Easy | Medium | Hard | Easy | Medium | Hard | Easy | Medium | Hard |  |
| **Single-Hop** | 2 | 4 | 1 | 4 | 2 | 1 | 0 | 3 | 3 | 19 |
| **Multi-Hop** | 1 | 3 | 1 | 4 | 2 | 0 | 0 | 2 | 3 | 17 |

It is interesting to note here that in the prompt provided to the LLM, for both the single-hop and multi-hop dataset generation tasks, the total number of questions specified to be generated (20 Q&As), along with the distribution of the question types to be generated is the same, i.e. 6 Factual, 7 True/False, and 7 Reasoning. For the single-hop dataset generation this instruction was followed more closely by the LLM as compared to that for the multi-hop dataset generation task. We believe this is due to the complexity, arising from the multi-step reasoning required to generate the questions for the multi-hop dataset. This compromise in performance is also seen in the ability of the LLM to generate answers for the multi-hop dataset. For the multi-hop dataset, the precision – a measure of the proportion of Q&A pairs generated that are from the context provided and correctly answered – is lower as compared to the single-hop dataset. In addition, the hallucination rate – a measure of the proportion of Q&A pairs generated out of context – is higher for the multi-hop dataset as compared to the single-hop dataset. While on the other hand, the accuracy – a measure of proportion of Q&As generated that are (i) from the context provided and correctly answered, and (ii) if out of context (i.e. hallucinated) then correctly identified as such — is higher for the multi-hop dataset as compared to the single-hop dataset; naturally the question then is why? This is because, the LLM while hallucinating more when generating the multi-hop questions, is also more 'cautious', and hence is able to correctly identify and correct course when generating the answers, i.e. the hallucation capture rate is much higher for the multi-hop dataset (84 %) *vs.* the single-hop dataset (22 %). Our hypothesis here is that since the multi-hop Q&A generation task is more complicated, the LLM approaches is it with more caution and is more careful when generating the answers. This hypothesis is further corroborated by the latency — which is a measure of the time required from the initiation of the request to the completion of the response — is found to be higher for the multi-hop dataset *vs.* the single-hop dataset for the same input provided.

A summary of the performance assessment of the single-hop and multi-hop Q&A datasets is given in **Table 2** while a more detailed, performance assessment by question type is shown in **Figure 6**. For the generation of the synthesis conditions task, the ratio of the number of **Y**s to the total number of outcomes (no. of **N**s + no. of **Y**s) was determined to be 0.794 for criterion 1 indicating that ~80 % of the time, the LLM was able to correctly extract all the synthesis conditions for a given material. The ratio of the number of **N**s to the total number of outcomes (no. of **N**s + no. of **Y**s) for criterion 2 was determined to be 0.893 indicating that ~90 % of the time, the LLM did not extract the experimental characterization. The obedience score was determined to be 0.708, indicating that ~70 % of the time the LLM followed both the instructions provided in the prompt. In total, the synthesis conditions generated for 238 DOIs were checked manually. The evaluations for all the datasets (including the single-hop and multi-hop datasets) is given in the supplementary files.

**Table 2: Performance assessment of the Single-Hop and Multi-Hop Q&A datasets.** The datasets are evaluated based on the following metrics: Accuracy = (TP+TN)/(TP+TN+FP+FN); Precision = TP/(TP+TN+FP+FN); Hallucination Rate = (TN+FN)/(TP+TN+FP+FN), and Hallucination Capture Rate = (TN)/(TN+FN). TP = True Positive; TN = True Negative, FP = False Positive, and FN = False Negative. For Accuracy, Precision, and Hallucination Capture Rate, higher values indicate better performance (1 indicates perfect performance), while for hallucination rate, lower values indicate better performance (0 indicates perfect performance). For the Single-hop dataset a total of 265 DOIs were evaluated while for the multi-hop dataset a total of 233 DOIs were evaluated.

|  | **Accuracy** | **Precision** | **Hallucination Rate** | **Hallucination Capture Rate** |
|---|---|---|---|---|
| **Single-Hop** | 0.948 | **0.943** | **0.028** | 0.217 |
| **Multi-Hop** | **0.983** | 0.934 | 0.055 | **0.841** |

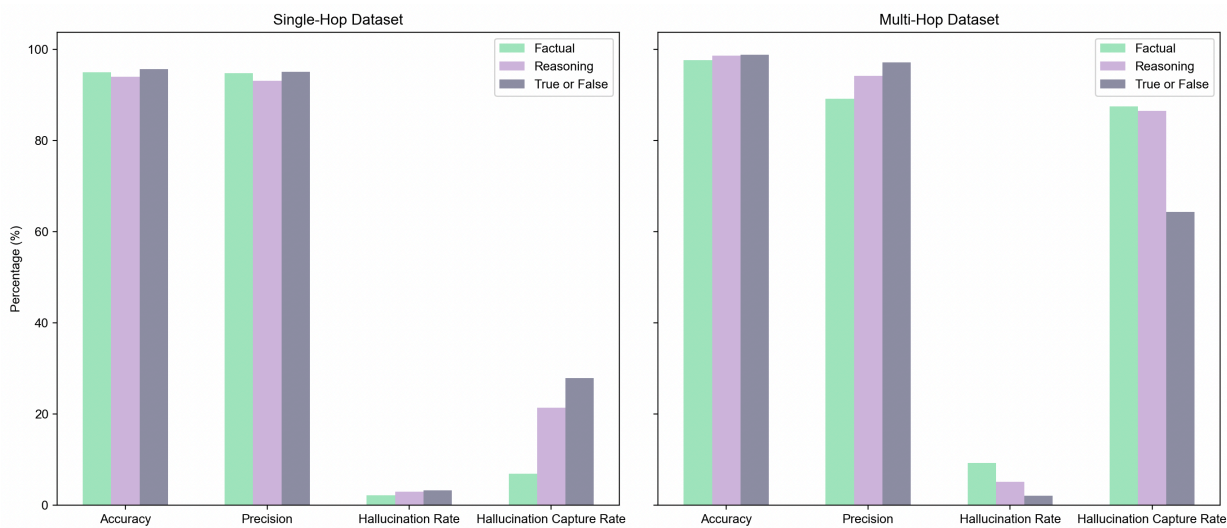

**Figure 6. Performance assessment by question type for both the Single-hop and Multi-hop Q&A datasets.** A comparison of the performance of the LLM as evaluated based on the accuracy, precision, hallucination rate, and hallucination capture rate for the single-hop dataset generation task (left) and the multi-hop dataset generation task (right).

**Concluding Remarks:**

In this study, we present a question-answering dataset specific to reticular chemistry. The dataset is generated automatically using an LLM, in this case GPT-4-Turbo. The LLM performs exceptionally well in generating both single-hop and multi-hop question-answer pairs with precision values of ~94 % indicating that majority of the time the Q&A pair generated is from the context provided (and not hallucinated). Interestingly, we find that when the task at hand is more complex for example, the task of generating a multi-hop Q&A pair, the LLM although hallucinates more is also more 'careful' in then evaluating the answers it generates, and therefore, we hypothesize is able to rectify its mistake, as indicated by an exceptionally high hallucination capture rate of 84 % for the multi-hop Q&A dataset. This eventually gave us a higher accuracy score for the multi-hop Q&A dataset.

The extraction and classification of synthesis conditions is far more complex given that the format of the synthesis conditions reported is different for each DOI. There is not a 'fixed' template or a given set of variables to follow for the LLM when extracting the synthesis conditions. The LLM exhibits an obedience score of ~70 % indicating that approximately in 7 out of 10 instances the LLM is able to both extract all the given set of synthesis conditions and also making sure that no experimental characterization data is included. While improving the performance of the LLM by developing more accurate models tuned for data extraction tasks is one way to address the low obedience score, another way would be to address the format of the data funnel that enters the LLM. In this case, we propose the development of a synthesis conditions information file (.sif) that similar to its counterpart the crystallographic information file (.cif) standardizes the reporting of synthesis conditions. We envision that the RetChemQA dataset will help (i) catalyze the development of large language models purpose built for reticular chemistry, and (ii) help democratize access to LLMs by enabling the development and application of automated prompt optimization frameworks, leading to an improvement in the reliability and accuracy of the outcomes generated by an LLM in the domain of reticular chemistry.

**References:**

[1] Q. Jin, B. Dhingra, Z. Liu, W. W. Cohen, and X. Lu, "PubMedQA: A Dataset for Biomedical Research Question Answering," *EMNLP-IJCNLP 2019 - 2019 Conference on Empirical Methods in Natural Language Processing and 9th International Joint Conference on Natural Language Processing, Proceedings of the Conference*, pp. 2567–2577, Sep. 2019, doi: 10.18653/v1/d19-1259.

[2] Z. Yang *et al.*, "HotpotQA: A Dataset for Diverse, Explainable Multi-hop Question Answering," *Proceedings of the 2018 Conference on Empirical Methods in Natural Language Processing, EMNLP 2018*, pp. 2369–2380, Sep. 2018, doi: 10.18653/v1/d18-1259.

[3] P. Rajpurkar, J. Zhang, K. Lopyrev, and P. Liang, "SQuAD: 100,000+ Questions for Machine Comprehension of Text," *EMNLP 2016 - Conference on Empirical Methods in Natural Language Processing, Proceedings*, pp. 2383–2392, Jun. 2016, doi: 10.18653/v1/d16-1264.

[4] P. Rajpurkar, R. Jia, and P. Liang, "Know What You Don't Know: Unanswerable Questions for SQuAD," *ACL 2018 - 56th Annual Meeting of the Association for Computational Linguistics, Proceedings of the Conference (Long Papers)*, vol. 2, pp. 784–789, Jun. 2018, doi: 10.18653/v1/p18-2124.

[5] O. M. Yaghi, "Reticular Chemistry in All Dimensions," *ACS Cent Sci*, vol. 5, no. 8, pp. 1295–1300, Aug. 2019, doi: 10.1021/ACSCENTSCI.9B00750/ASSET/IMAGES/MEDIUM/OC9B00750_0003.GIF.

[6] O. M. Yaghi, "Reticular Chemistry—Construction, Properties, and Precision Reactions of Frameworks," *J Am Chem Soc*, vol. 138, no. 48, pp. 15507–15509, Dec. 2016, doi: 10.1021/JACS.6B11821.

[7] W. X. Zhao *et al.*, "A Survey of Large Language Models," Mar. 2023, Accessed: Apr. 23, 2024. [Online]. Available: https://arxiv.org/abs/2303.18223v13

[8] O. Khattab *et al.*, "DSPy: Compiling Declarative Language Model Calls into Self-Improving Pipelines," Oct. 2023, Accessed: Apr. 23, 2024. [Online]. Available: https://arxiv.org/abs/2310.03714v1

[9] B. Wang, A. P. Côté, H. Furukawa, M. O'Keeffe, and O. M. Yaghi, "Colossal cages in zeolitic imidazolate frameworks as selective carbon dioxide reservoirs," *Nature 2008 453:7192*, vol. 453, no. 7192, pp. 207–211, May 2008, doi: 10.1038/nature06900.

[10] K. S. Park *et al.*, "Exceptional chemical and thermal stability of zeolitic imidazolate frameworks," *Proc Natl Acad Sci U S A*, vol. 103, no. 27, pp. 10186–10191, Jul. 2006, doi: 10.1073/PNAS.0602439103/SUPPL_FILE/02439SUPPAPPENDIX.PDF.

[11] A. Li, R. B. Perez, S. Wiggin, S. C. Ward, P. A. Wood, and D. Fairen-Jimenez, "The launch of a freely accessible MOF CIF collection from the CSD," *Matter*, vol. 4, no. 4, pp. 1105–1106, Apr. 2021, doi: 10.1016/j.matt.2021.03.006.

**Author Contributions:** N.R. C.B. J.C. and O.Y. conceived the idea and drafted the outline. N.R. wrote the initial draft of the manuscript, including the design of the figures. K.Y. led the evaluation of the multi-hop dataset. M.B. together with J.A.J. led the evaluation of the single-hop dataset. L.H. led the evaluation of the synthesis conditions dataset. A.S. made sure that the data scraping followed the guidelines and permissions as outlined in the agreement between the UC Berkeley Library and the publishers. All authors contributed to the review and editing of the final manuscript.

**Acknowledgements:** N.R. acknowledges the Bakar Institute of Digital Materials for the Planet (BIDMaP) Emerging Scholars Program for the funding that supports this work.

**Competing interests**: All authors declare no competing interests.

Supplementary Information

*for*

# Single and Multi-Hop Question-Answering Datasets for Reticular Chemistry with GPT-4-Turbo

Nakul Rampal[1,2,3], Kaiyu Wang[1,2], Matthew Burigana[1,2], Lingxiang Hou[1,2], Juri Al-Johani [3,4], Anna Sackmann[5], Hanan S. Murayshid[8], Walaa Abdullah Al-Sumari[8], Arwa M. Al-Abdulkarim[8], Nahla Eid Al-Hazmi[8,9], Majed O. Al-Awad[8], Christian Borgs[3,4,*], Jennifer T. Chayes[3,4,6,7,*], Omar M. Yaghi[1,2,3,10,*]

[1]Department of Chemistry, University of California, Berkeley, California 94720, United States

[2]Kavli Energy Nanoscience Institute, University of California, Berkeley, California 94720, United States

[3]Bakar Institute of Digital Materials for the Planet, College of Computing, Data Science, and Society, University of California, Berkeley, California 94720, United States

[4]Department of Electrical Engineering and Computer Sciences, University of California, Berkeley, California 94720, United States

[5]Data Services Librarian, University of California, Berkeley Library, United States

[6]Department of Mathematics, University of California, Berkeley, California 94720, United States

[7]Department of Statistics, University of California, Berkeley, California 94720, United States

[8]Artificial Intelligence & Robotics Institute, Economies of the Future Sector, King Abdulaziz City for Science and Technology (KACST), Riyadh 12354, Saudi Arabia

[9]Hydrogen Technologies Institute, King Abdulaziz City for Science and Technology, P.O. Box 6086, Riyadh 11442, Saudi Arabia

[10]KACST−UC Berkeley Center of Excellence for Nanomaterials for Clean Energy Applications, King Abdulaziz City for Science and Technology, Riyadh 11442, Saudi Arabia

*Email: borgs@berkeley.edu, jchayes@berkeley.edu, yaghi@berkeley.edu

**Synthesis Conditions**

```
"System": " You are a synthesis condition
classification agent. You are required to go through
the given text and identify the synthesis conditions
for each and every material given in the paper (both
MS and SI, if available). There may be information
about the synthesis condition of more than one
material. Please make sure to separate these
materials when generating the .json file. For each
material try to classify the conditions under
different labels such as temperature, solvents, the
amount of each solvent (this is important),
equipment, chemicals used, time, washing method,
drying method, yield, etc. This is not at an
exhaustive list of labels, please feel free to add
more labels as required. Some synthesis conditions
may involve multiple steps, please take that into
account. Please do not include any experimental
characterization data such as those from Powder X-Ray
Diffraction (PXRD), Infrared (IR) spectra, ,
adsorption isotherms, thermogravimetric analysis
(TGA), nuclear magnetic resonance (NMR) experiments
etc. (this is not an exhaustive list and there may be
other characterization techniques) including
information about its properties. I reiterate, please
do not include any experimental characterization data
"

"User": "Generate a machine readable .json file
containing the synthesis conditions for the following
text:{combined_text}."
```

**Figure S1. Prompt used to extract the synthesis conditions from a paper.** The "`system`" role provides the high-level instructions, while the "`user`" role provides the query. The "`combined_text`" variable holds all the text information contained in the MS and SI (where available). This information is provided as part of the prompt to GPT-4-Turbo.

**Algorithm 1** Dataset Generation Algorithm
---
1: **Input:** Directory path *doi_dir* containing document files, LLM model *llm_engine*, maximum token limit *max_tokens*, tokenizer *tokenizer*
2: Initialize *start_time* with current time
3: Initialize *combined_text* and *tokens* to empty
4: **for** each *publisher* in directory *base_path* **do**
5:     **for** each *doi* in directory *publisher_path* **do**
6:         **if** output directory for *doi* does not exist **then**
7:             *combined_text* ← PROCESSDOI(*doi_path*)
8:             *tokens* ← TOKENIZE(*combined_text*)
9:             **if** length of *tokens* exceeds *max_tokens* **then**
10:                 PROCESSTOKENS(*tokens*, *llm_engine*, *temperature*, *start_time*, *publisher*, *doi*, *home_dir*)
11:             **end if**
12:         **end if**
13:     **end for**
14: **end for**
15: **if** remaining *tokens* exist **then**
16:     PROCESSTOKENS(*tokens*, *llm_engine*, *temperature*, *start_time*, *publisher*, *doi*, *home_dir*)
17: **end if**
18: **function** TOKENIZE(*text*)
19:     *tokens* ← convert *text* into a list of tokens using *tokenizer*
20:     **return** *tokens*
21: **end function**
22: **function** PROCESSDOI(*doi_dir*)
23:     *combined_text* ← empty string
24:     **for** each file in *doi_dir* **do**
25:         *text* ← PROCESSFILE(file)
26:         *combined_text* ← *combined_text* + *text*
27:     **end for**
28:     **return** *combined_text*
29: **end function**
30: **function** PROCESSFILE(*file_path*)
31:     *ext* ← file extension from *file_path*
32:     **if** *ext* is '.pdf' **then**
33:         **return** EXTRACTTEXTFROMPDF(*file_path*)
34:     **else if** *ext* is '.docx' or *ext* is '.doc' **then**
35:         **return** PROCESSDOCX(*file_path*)
36:     **else if** *ext* is '.xml' **then**
37:         **return** PROCESSXML(*file_path*)
38:     **else if** *ext* is '.xhtml' **then**
39:         **return** PROCESSXHTML(*file_path*)
40:     **end if**
41: **end function**
42: **function** PROCESSTOKENS(*tokens*, *llm_engine*, *temperature*, *start_time*, *publisher*, *doi*, *home_dir*)
43:     Generate multi-hop Q&A using *llm_engine* and *tokens*
44:     Store results in *home_dir* with appropriate *file_name*
45: **end function**

**Figure S2. Algorithm used to run the data set generation.** The algorithm described above is used to read files in the folder *publisher_path* in different formats such as (.docx, .xml, .pdf, and .xhtml) and then tokenize them to be then passed to the LLM for processing. Once the output has been generated, it is stored in the *home_dir*. This is a detailed version of the dataset generation workflow shown in **Figure 3.** The variables are as follows: *llm_engine* = GPT-4-Turbo (gpt-4-0125-preview), *max_tokens* = 128000, and *tokenizer* = GPT2, and *temperature* = 0.

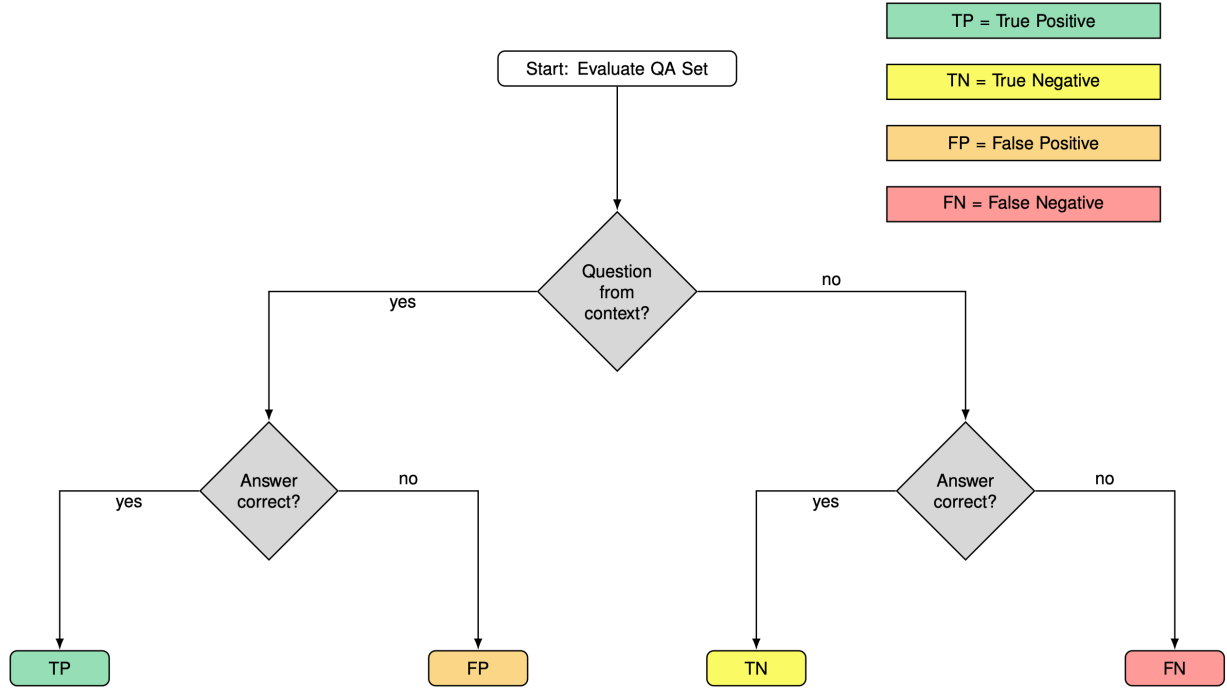

**Figure S3. Evaluation criteria for each Q&A in the single hop and multi-hop datasets.** In the first step, the question is screened on whether it has been generated from the context provided in the prompt to the LLM; If yes, then we check if the question is answered correctly or not, partially/incorrectly answered questions are classified as 'FP' and perfectly answered questions are classified as 'TP'; If not, we check if the answer to the 'hallucinated' question generated is correct or not, if it is correct, it is classified as 'TN', and otherwise 'FN'. Questions that have been incorrectly categorized, or, if the LLM has generated an incorrect question type, for example, the prompt explicitly states that multi-hop question-answer pairs should be generated but instead it generates single-hop question-answer pairs, or vice-versa, are classified as 'out of context' and evaluated as such.

# Single-hop Q&A

### True Positive (TP) — Output

**"Question":** "Based on the atomic parameters provided, how many hydrogen atoms' parameters are not refined?"

**"Answer":** "10"

**"Difficulty":** "Medium"

**"Type":** "Reasoning"

---

Questions are classified as 'TP' when (i) the question is from the context provided, and (ii) the entire "Answer" is correct.

**DOI:** 10.1021/jacs.00211a017

### False Positive (FP) — Output

**"Question":** "What is the general formula for the synthesis of compounds 1–5?"

**"Answer":** "3 H2L + EuCl3 x 6 H2O –> EuL3 3– + 3 Cl– 6H2O + 6 H+ –> EuL3 3– + 3 Cl– + 6 H+ + 2 BaCl2 x 2 H2O –> Ba2ClEuL3 x 10 H2O + 6HCl"

**"Difficulty":** "Medium"

**"Type":** "Factual"

---

Questions are classified as 'FP' when (i) the question is from the context, and (ii) the answer is *incomplete* or wrong. In the example above, the answer is incorrect.

**DOI:** 10.1021/ja060666e

### True Negative (TN) — Output

**"Question":** "Does the crystal structure of 1 change significantly under high pressure"

**"Answer":** "The text does not provide specific details on structural changes under high pressure, only the resistivity behaviour."

**"Difficulty":** "Hard"

**"Type":** "True or False"

---

Questions are classified as 'TN' when (i) the question is not from the context provided, but (ii) answered correctly, and (iii) if the questions have been incorrectly classified or a question of wrong type has been generated. In the example above, the question is out of context but has been answered correctly.

**DOI:** 10.1002/anie.200600553

### False Negative (FN) — Output

**"Question":** "What is the role of pore size, ligand functionalization, and exposed metal sites in high capacity hydrogen adsorption in Cu(II) tetracarboxylate framework materials?"

**"Answer":** "The role of pore size, ligand functionalization, and exposed metal sites is crucial for high capacity hydrogen adsorption in Cu(II) tetracarboxylate framework materials as they significantly enhance hydrogen uptake."

**"Difficulty":** "Hard"

**"Type":** "Reasoning"

---

Questions are classified as 'FN' when (i) the question is not from the context provided, and (ii) the "Answer" is incorrect.

**DOI:** 10.1021/jacs.0c11930

**Figure S4.** Examples of questions classified as TP, FP, TN, and FN in the single-hop dataset.

# Multi-hop Q&A

### True Positive (TP) — Output

**"Question":** "How does the adsorption behaviour of CUK-1 and CUK-2 differ for $H_2$ at 77 K?"

**"Answer":** "CUK-1 shows higher H2 uptake of 183 $cm^3$ $g^{-1}$ at 760 Torr compared to CUK-2's uptake of 75 mL $g^{-1}$ at 760 Torr."

**"Difficulty":** "Medium"

**"Type":** "Factual"

Questions are classified as 'TP' when (i) the question is from the context provided, and (ii) the entire "Answer" is correct.

**DOI:** 10.1002/anie.200601627

### False Positive (FP) — Output

**"Question":** "How does the coordination polymer's response to UV light vary with crystal shape and dimensions?"

**"Answer":** "The type of photomechanical response, such as bending, bursting, or, jumping, varies depending on the crystal's shape and dimensions."

**"Difficulty":** "Hard"

**"Type":** "Factual"

Questions are classified as 'FP' when (i) the question is from the context, and (ii) the answer is *incomplete* or wrong. In the example above, the answer is incomplete.

**DOI:** 10.1002/anie.202306048

### True Negative (TN) — Output

**"Question":** "What chemical reaction is triggered by UV light in the coordination polymer?"

**"Answer":** "[2+2] cycloaddition reaction"

**"Difficulty":** "Easy"

**"Type":** "Reasoning"

Questions are classified as 'TN' when (i) the question is not from the context provided, but (ii) answered correctly, and (iii) if the questions have been incorrectly classified or a question of wrong type has been generated. In the example above, the question has been incorrectly classified.

**DOI:** 10.1002/anie.202306048

### False Negative (FN) — Output

**"Question":** "What is the significance of the crystallographic space groups C2/c and C2221 for CUK-1 and CUK-2 respectively?"

**"Answer":** "These space groups indicate the unique crystal structures of CUK-1 and CUK-2, which are crucial for their gas adsorption properties and stability."

**"Difficulty":** "Hard"

**"Type":** "Reasoning"

Questions are classified as 'FN' when (i) the question is not from the context provided, and (ii) the "Answer" is incorrect.

**DOI:** 10.1002/anie.200601627

**Figure S5.** Examples of questions classified as TP, FP, TN, and FN in the multi-hop dataset.

**Table S1. List of acronyms and their full forms.**

| Acronym | Full form |
|---------|-----------|
| NAS | National Academy of Sciences |
| ACS | American Chemical Society |
| RSC | Royal Society of Chemistry |
| T&F | Taylor & Francis |
| AAAS | American Association for the Advancement of Science |
| AIP | American Institute of Physics |
| APS | American Physical Society |
| CCS | Chinese Chemical Society |
| IOP | Institute of Physics |
| IUCr | International Union of Crystallography |

**Table S2. List of Publishers, and the associated journals, along with the number of publications from each journal used for the dataset generation.**

| Publisher | Journal | No. of publications |
|---|---|---|
| Elsevier | Chem | 25 |
| | Chemical Engineering Journal | 57 |
| RSC | Chemical Science | 22 |
| | Dalton Transactions | 63 |
| | Chemical Communications | 130 |
| Nature Publishing Group | Nature | 26 |
| | Nature Synthesis | 2 |
| | Nature Catalysis | 3 |
| | Nature Sustainability | 1 |
| | Nature Materials | 15 |
| | Nature Energy | 3 |
| | Nature Chemistry | 44 |
| | Nature Biomedical Engineering | 1 |
| | Nature Communications | 124 |
| | Nature Photonics | 1 |
| ACS | Journal of the American Chemical Society | 1,283 |
| AAAS | Science Advances | 12 |
| | Science | 34 |
| Wiley | Angewandte Chemie | 593 |
| | Advanced Functional Materials | 39 |
| | Advanced Materials | 21 |
| NAS | Proceedings of the National Academy of Sciences | 10 |
| CCS | CCS Chemistry | 10 |
| AIP | APL Materials | 1 |
| | Journal of Applied Physics | 3 |
| | Structural Dynamics | 1 |
| APS | Physical Review B | 4 |
| | Physical Review Letters | 2 |